\documentclass[runningheads]{llncs}
\usepackage{graphicx}
\usepackage{ amssymb }
\usepackage{mathtools}
\usepackage{multirow}

\DeclarePairedDelimiter\abs{\lvert}{\rvert}%
\DeclarePairedDelimiter\norm{\lVert}{\rVert}%

\makeatletter
\let\oldabs\abs
\def\abs{\@ifstar{\oldabs}{\oldabs*}}
\let\oldnorm\norm
\def\norm{\@ifstar{\oldnorm}{\oldnorm*}}
\makeatother

\begin{document}
%
%\title{GANs for Simultaneous Synthesis of Anatomic and Aptw MRIs using Lesion Mask}

\title{Lesion Mask-based Simultaneous\\ Synthesis of Anatomic and Molecular \\MR Images using a GAN}
\titlerunning{Lesion Mask-based Synthesis of Anatomic and Molecular MRI}
% If the paper title is too long for the running head, you can set
% an abbreviated paper title here
%
\author{Pengfei Guo\inst{1}, Puyang Wang\inst{2}, Jinyuan Zhou\inst{3}, \\ Vishal M. Patel\inst{1,2}, Shanshan Jiang\inst{3}}
%index{Guo, Pengfei}
%index{Wang, Puyang}
%index{Zhou, Jinyuan}
%index{Patel, Vishal}
%index{Jiang, Shanshan}

\authorrunning{P. Guo et al.}
% First names are abbreviated in the running head.
% If there are more than two authors, 'et al.' is used.
%
\institute{Department of Computer Science, Johns Hopkins University, MD, USA\\
\and
Department of Electrical and Computer Engineering, Johns Hopkins University, MD, USA\\
\and
Department of Radiology, Johns Hopkins University, Baltimore, MD, USA\\ \email{sjiang21@jhmi.edu}}
%\\ \email{**@******.***}}
%
\maketitle              % typeset the header of the contribution
\begin{abstract}
Data-driven automatic approaches have demonstrated their great potential in resolving various clinical diagnostic dilemmas for patients with malignant gliomas in neuro-oncology with the help of conventional and advanced molecular MR images. However, the lack of sufficient annotated MRI data has vastly impeded the development of such automatic methods.  Conventional data augmentation approaches, including flipping, scaling, rotation, and distortion are not capable  of generating data with diverse image content. In this paper, we propose a method, called synthesis of anatomic and molecular MR images network (SAMR), which can simultaneously synthesize data from arbitrary manipulated lesion information on multiple anatomic and molecular MRI sequences, including T1-weighted ($T_1$w), gadolinium  enhanced $T_1$w (Gd-$T_1$w),  T2-weighted ($T_2$w), fluid-attenuated inversion recovery ($FLAIR$), and amide proton transfer‐weighted ($APT$w). The proposed framework consists of a stretch-out up-sampling module, a brain atlas encoder, a segmentation consistency module, and multi-scale label-wise  discriminators. Extensive experiments on real clinical data demonstrate that the proposed model can perform significantly better than the state-of-the-art synthesis methods.

\end{abstract}
\section{Introduction}
Malignant gliomas, such as glioblastoma (GBM), remain one of the most aggressive forms of primary brain tumor in adults. The median survival of patients with glioblastomas is only 12 to 15 months with aggressive treatment \cite{au2}. For the clinical management in patients who finish surgery and chemoradiation, the treatment responsiveness assessment is relied on the pathological evaluations \cite{au1}. In recent years, deep convolutional neural network (CNN) based medical image analysis methods have shown to produce significant improvements over the conventional methods \cite{au3,au22}. However, a large amount of data with rich diversity is required for training effective CNNs models, which is usually unavailable for medical image analysis. Furthermore, lesion annotations and image prepossessing (e.g. co-registration) are labor-intensive, time-consuming and expensive, since expert radiologists are required to label and verify the data. While deploying conventional data augmentations, such as rotation, flipping, random cropping, and distortion, during training partly mitigates such issues, the performance of CNN models still suffer from the limited diversity of the dataset \cite{au4}. In this paper, we propose a generative network which can simultaneously synthesize meaningful high quality $T_1$w, Gd-$T_1$w, $T_2$w, $FLAIR$, and $APT$w MRI sequences from input lesion mask.  In particular, $APT$w is a novel molecular MRI technique, which yields a reliable marker for treatment responsiveness assessment for patients with post-treatment malignant gliomas \cite{au5,au6}.\\
\indent Recently Goodfellow et al. \cite{au7} proposed generative adversarial network (GAN) which has been shown to synthesize photo-realistic images. Isola et al. \cite{au8} and Wang et al. \cite{au9} applied GAN under the conditional settings and achieved impressive results on image-to-image translation tasks. When considering the generative models for MRI synthesis alone, several methods have been proposed in the literature. Nguyen et al. \cite{au10} and  Chartsias et al. \cite{au12} proposed CNN-based architectures integrating intensity features from images to synthesize cross-modality MR images. However, their inputs are existing MRI modalities and the diversity of the synthesized images is limited by the training images. Cordier et al. \cite{au13} used a generative model for multi-modal MR images with brain tumors from a single label map. However, the input label map contains detailed brain anatomy and the method is not capable of producing manipulated outputs. Shin et al. \cite{au14} adopted pix2pix \cite{au8} to transfer brain anatomy and lesion segmentation maps to multi-modal MR images with brain tumors.  However, it requires to train an extra segmentation network that provides white matter, gray matter, and cerebrospinal fluid (CSF) masks as partial input of synthesis network. Moreover, it is only demonstrated to synthesize anatomical MRI sequences.  In this paper, a novel generative model is proposed that can take arbitrarily manipulated lesion mask as input facilitated by brain atlas generated from training data to simultaneously synthesize a diverse set of anatomical and molecular MR images.\\  
 \indent To summarize, the following are our key contributions: \textbf{1.} A novel conditional GAN-based model is proposed to synthesize meaningful high quality multi-modal anatomic and molecular MR images with controllable lesion information. \textbf{2.} Multi-scale label-wise discriminators are developed to provide specific supervision on the region of interest (ROI). \textbf{3.}  Extensive experiments are conducted and comparisons are performed against several recent state-of-the-art image synthesis approaches. Furthermore, an ablation study is conducted to demonstrate the improvements obtained by various components of the proposed method.

%\textbf{1.} An automatic, low-cost method is proposed to generate data with diverse content that can be added to the training dataset (e.g. changing lesion locations, scaling the tumor’s size, and reassembling lesions between patients). 
\section{Methodology}
\vskip-10pt	
\begin{figure}
\centering
\includegraphics[width=\textwidth]{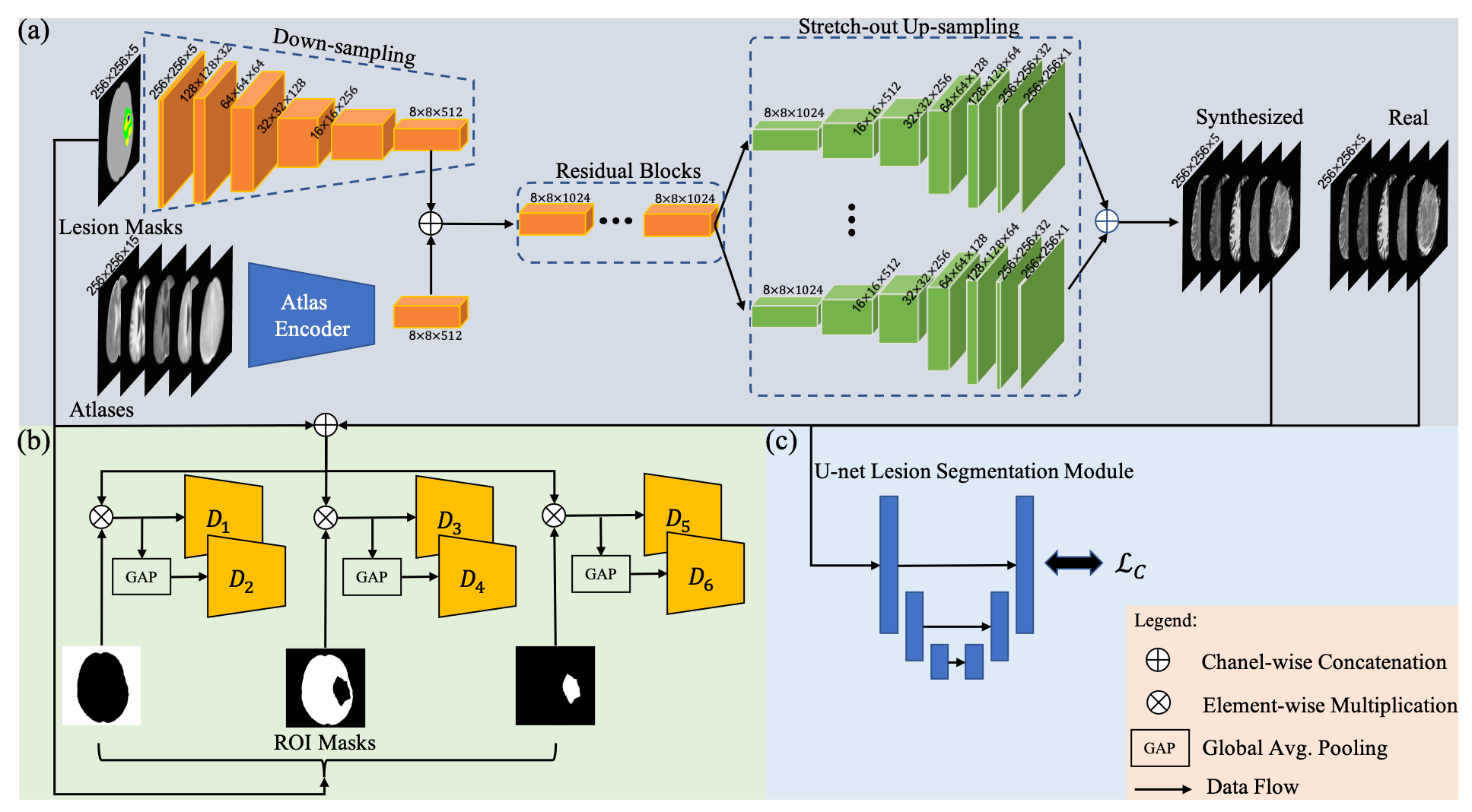}
\vskip-13pt	\caption{An overview of the proposed network.  (a) Generator network. (b) Multi-scale Label-wise discriminators. Global averaging pooling is used to create the factor of 2 down-sampling input. (c) U-net based lesion segmentation module. We denote lesion shape consistency loss as $\mathcal{L}_{C}$. \label{fig1}}
\vskip-20pt	
\end{figure}
Fig.~\ref{fig1} gives an overview of the proposed framework. Incorporating multi-scale label-wise discriminators and shape consistency-based optimization, the generator aims to produce meaningful high-quality anatomical and molecular MR images with diverse and controllable lesion information. In what follows, we describe different parts of the network in detail.\\ 

%While applying 3D convolution operations might reflect the reality of data, the output of proposed method is multi-modal MRI image slices, since voxel size between anatomical and molecular MRI in axial direction is significantly different and re-sampling to isotropic resolution can severely degrade image quality. Detailed Imaging parameters is shown in Sect. \ref{2.3}. In the following subsections, we will introduce the architecture of the proposed generator and discriminators, lesion consistency optimization, and data acquisition.
%\subsection{Multi-modal MRI Sequence  Generation }

\noindent {\bf{Multi-modal MRI Sequence  Generation. }}  Our generator architecture is inspired by the models proposed by Johnson et al. \cite{au15} and Wang et al. \cite{au9}. The generator network, consists of four components (see Fig.~\ref{fig1}(a): a down-sampling module, an atlas encoder, a set of residual blocks, and a stretch-out up-sampling module. A lesion segmentation map of size 256 $\times$ 256 $\times$ 5, containing 5 labels: background, normal brain, edema, cavity caused by surgery, and tumor, is passed through the down-sampling module to get a latent feature map. The corresponding multi-model atlas of size 256 $\times$ 256 $\times$ 15 (details of atlas generation are provided in Section 3) is passed through an atlas encoder to get another latent feature map. Then, the two latent feature maps are concatenated and are passed through residual blocks and stretch-out up-sampling module to synthesize multi-model MRI slices of size 256 $\times$ 256 $\times$ 5. \\
\indent The down-sampling module consists of a fully-convolutional module with 6 layers.  We set the kernel size and stride equal to 7 and 1, respectively for the first layer. For down-sampling, instead of using maximum-pooling, the stride of other 5 layers is set equal to 2. Rectified Linear Unit (ReLu) activation and batch normalization are sequentially added after each layer. The atlas encoder has the same network architecture but the number of channels in the first convolutional layer is modified to match the input size of the multi-model atlas input. The depth of the network is increased by a set of residual blocks, which is proposed to learn better transformation functions and representations through a deeper perception \cite{au4}. The stretch-out up-sampling module contains 5 similar sub-modules designed to utilize the same latent representations from residual blocks and perform customized synthesis for each sequence. Each sub-module contains one residual learning block and a symmetric architecture with a down-sampling module.  All convolutional layers are replaced by transposed convolutional layers for up-sampling. The synthesized multi-model MR images are produced from each sub-model.
		
%\subsection{Multi-scale Label-wise Discriminators and Optimization }\label{2.2}
%\subsubsection{Discriminators.}  
\noindent {\bf{Multi-scale Label-wise Discriminators. }}  In order to efficiently achieve large receptive field in discriminators, we adopt multi-scale PatchGAN discriminators \cite{au8}, which have identical network architectures but take multi-scale inputs \cite{au9}. Conventional discriminators operate on the combination of images and conditional information to distinguish between real and synthesized images. However, optimizing generator to produce realistic images in each ROI cannot be guaranteed by discriminating on holistic images. To address this issue, we propose label-wise discriminators. Based on the radiographic features, original lesion segmentation masks are reorganized into 3 ROIs, including background, normal brain, and lesion. Specifically, the input of each discriminator is the ROI-masked combination of lesion segmentation maps and images. Since proposed discriminators are in a multi-scale setting,  for each ROI there are 2 discriminators that operate on original and a down-sampled scales (factor of 2). Thus, there are in total 6 discriminators for 3 ROIs and we refer to these set of discriminators as $\mathbb{D}=\{D_1, D_2, D_3, D_4, D_5, D_6\}$. In particular, \{$D_1$,$D_2$\},\{$D_3$,$D_4$\}, and \{$D_5$,$D_6$\} operate on original and down-sampled versions of background, normal brain, and lesion, respectively. An overview of the proposed discriminators is shown in Fig.~\ref{fig1}(b). The objective function for a specific discriminator  $\mathcal{L}_{GAN}(G,D_k)$ is as follows:
\setlength{\belowdisplayskip}{0pt} \setlength{\belowdisplayshortskip}{0pt}
\setlength{\abovedisplayskip}{0pt} \setlength{\abovedisplayshortskip}{0pt}
\begin{equation} \label{eq:3}
\mathcal{L}_{GAN}(G,D_k)=\mathbb{E}_{(\hat{x},\hat{y})}[\log D_{k}(\hat{x},\hat{y})] + \mathbb{E}_{\hat{x}}[\log (1- D_{k}(\hat{x}, \hat{G}(x)))],
\end{equation}
where $x$ and $y$ are paired original lesion segmentation masks and real multi-model MR images, respectively. Here, $\hat{x} \triangleq c_k\odot x$, $\hat{y} \triangleq c_k\odot y$, and $\hat{G}(x) \triangleq c_k\odot G(x)$, where  $\odot$ denotes element-wise multiplication and $c_k$ corresponds to the ROI mask.  For simplicity, we omit the down-sampling operation in this equation.\\
%$\hat{G}(x)$ are the synthezied multi-model MR images and $c_k$ is the corresponding ROI mask. We denote $\odot$ as element-wise multiplication and $\mathbb{E}_{\hat{x}}$ is defined to be $\mathbb{E}_{\hat{x} \sim p_{data}(\hat{x})}$. $\hat{x}$, $\hat{y}$, and $\hat{G}(x)$ stand for the lesion segmentation maps, real, and synthesized images corresponding to a specific ROI (i.e. $\hat{x} \triangleq c_k\odot x$, $\hat{y} \triangleq c_k\odot y$, and $\hat{G}(x) \triangleq c_k\odot G(x)$). 

%\subsubsection{Multi-task Optimization.}
\noindent {\bf{Multi-task Optimization. }}
A multi-task loss is designed to train the generator and discriminators in an adversarial setting. Instead of only using the conventional adversarial loss $\mathcal{L}_{GAN}$, we also adopt a feature matching loss $\mathcal{L}_{FM}$ \cite{au9} to stabilize training, which optimizes generator to match these intermediate representations from the real and the synthesized images in multiple layers of the discriminators. For a specific discriminator, $\mathcal{L}_{FM}(G,D_k)$ is defined as follows:  
%\begin{equation} \label{eq:4}
%\mathcal{L}_{FM}(G,D_k)= \sum_{i}^{T}  \frac{1}{N_{i}} \left[\|D_{k}^{(i)}(x,c\odot y)- D_{k}^{(i)}(x,c \odot G(x)\|_{2}^{2}\right],
%\end{equation}
\begin{equation} \label{eq:4}
\mathcal{L}_{FM}(G,D_k)= \sum_{i}^{T}  \frac{1}{N_{i}} \left[\|D_{k}^{(i)}(\hat{x},\hat{y})- D_{k}^{(i)}(\hat{x}, \hat{G}(x)\|_{2}^{2}\right],
\end{equation}
where $D_{k}^{(i)} $ denotes the $i$th layer of the discriminator $D_{k}$, $T$ is the total number of layers in $D_{k}$ and $N_i$ is the number of elements  in the $i$th layer. If we perform lesion segmentation on images, it is worth to note that there is a consistent relation between the prediction and the real one serving as input for the generator. Lesion labels are usually occluded with each other and brain anatomic structure, which causes ambiguity for synthesizing realistic MR images. To tackle this problem, we propose a lesion shape consistency loss $\mathcal{L}_{C}$ by adding a  U-net \cite{au11} segmentation module (Fig.~\ref{fig1}(c)), which regularizes the  generator to obey this consistency relation. We adopt Generalized Dice Loss ($GDL$) \cite{au16} to measure the difference between predicted and real segmentation maps and is defined as follows:
\begin{equation} \label{eq:7}
GDL(R,S) =  1 - \frac{2\sum_i^N r_i s_i }{\sum_i^N r_i + \sum_i^N s_i },
\end{equation}
where $R$ denotes the ground truth and $S$ is the segmentation result. $r_i$ and $s_i$ represent the ground truth and predicted probabilistic maps  at each pixel $i$, respectively. $N$ is the total number of pixels. The lesion shape consistency loss $\mathcal{L}_{C}$ is then defined as follows:       
\begin{equation} \label{eq:5}
\mathcal{L}_{C}(x, U(G(x)),U(y)) = GDL(x,U(y)) + GDL(x,U(G(x))),
\end{equation}
where $U(y)$ and $U(G(x))$ represent the predicted probabilistic maps by taking $y$ and $G(x)$ as inputs in the segmentation module, respectively. The proposed final multi-task loss function for the generator is defined as:
\begin{equation} \label{eq:6}
\mathcal{L}= \sum_{k=1}^{6} \mathcal{L}_{GAN} (G,D_k )  +\lambda_{1}\sum_{k=1}^{6}\mathcal{L}_{FM}(G,D_k )+\lambda_{2}\mathcal{L}_{C}(x, U(G(x)),U(y)),
\end{equation}
%\begin{equation} \label{eq:6}
%\min\limits_{G}(\max\limits_{\mathbb{D}} \sum_{k=1}^{C} \mathcal{L}_{GAN}(G,D_k ) +  \lambda_{1}\sum_{k=1}^{C} \mathcal{L}_{FM}(G,D_k)) + \min\lambda_{2}\mathcal{L}_{Consistency}(x,G(x),y),
%\end{equation}
where $\lambda_{1}$ and $\lambda_{2}$ two parameters that control the importance of each loss.
\section{Experiments and  Evaluations}
\noindent {\bf{Data Acquisition and Implementation Details. }} 90 postsurgical patients were involved in this study. MRI scans were acquired by a 3T human MRI scanner (Achieva; Philips Medical Systems). Anatomic sequences of size 512 $\times$  512 $\times$ 150 voxels and Molecular $APT$w sequence of size 256 $\times$  256 $\times$ 15 voxels were collected. Detailed imaging parameters and preprocessing pipeline can be found in supplementary material. After preprocessing, the final volume size of each sequence is 256 $\times$  256 $\times$ 15. Expert neuroradiologist manually annotated five labels (i.e. background, normal brain, edema, cavity and tumor) for each patient. Then, a multivariate template construction tool \cite{au21} was used to create the group average for each sequence (atlas). 1350 instances with the size of 256 $\times$ 256 $\times$  5 were extracted from the volumetric data, where the 5 corresponds to five MR sequences. For every instance, the one corresponding atlas slice and two adjacent (in axial direction) atlas slices were extracted to provide human brain anatomy prior. We split these instances randomly into 1080 (80\%) for training and 270 (20\%)  for testing on the patient level.  Data collection and processing are approved by the Institutional Review Board.\\
\indent The synthesis model was trained based on the final objective function Eq. (\ref{eq:6}) using the Adam optimizer \cite{au21}. $\lambda_{1}$ and $\lambda_{2} $ in Eq. (\ref{eq:6}) were set equal to 5 and 1, respectively. Hyperparameters are set as follows: constant learning rate of 2 $\times 10^{-4}$ for the first 250 epochs then linearly decaying to 0; 500 maximum epochs; batch size of 8. For evaluating the effectiveness of the synthesized MRI sequences on data augmentation, we leveraged U-net \cite{au11} to train lesion segmentation models. U-net \cite{au11} was trained by the Adam optimizer \cite{au21}. Hyperparameters are set as follows: constant learning rate of 2 $\times 10^{-4}$ for the first 100 epochs then linearly decaying to 0; 200 maximum epochs; batch size of 16. In the segmentation training, all the synthesized data was produced by randomly manipulated lesion masks. For evaluation, we always keep 20\%  of data unseen for both of the synthesis and segmentation models.\\

\begin{figure}[htp!]
	\centering
		\vskip-30pt	
	\includegraphics[scale=0.3]{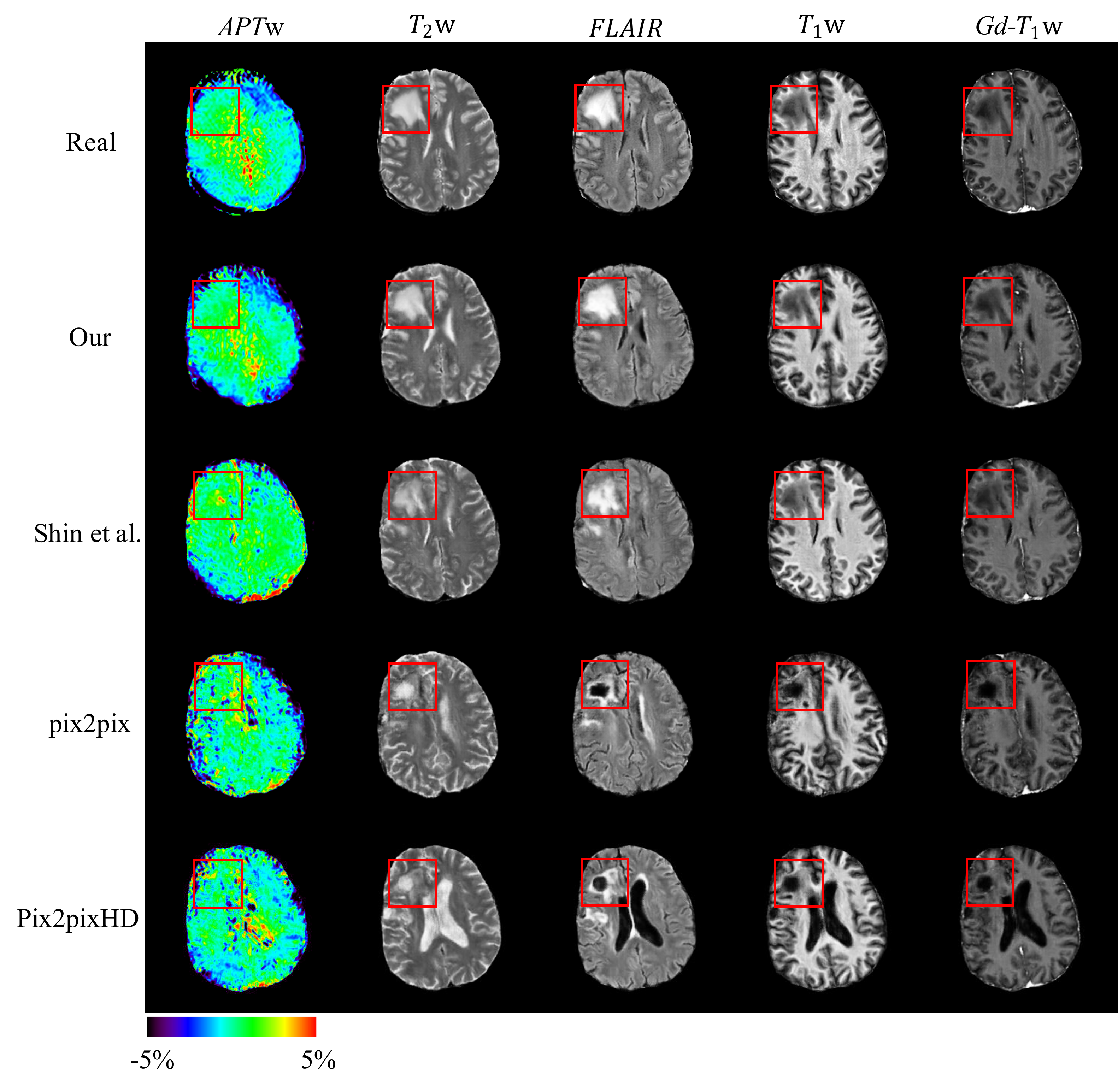}
	\vskip-15pt	\caption{Qualitative comparison of different methods.  The same lesion mask is used to synthesize images from different methods. Red boxes indicate the lesion region.} \label{fig3}
	\vskip-15pt	
\end{figure}
\begin{figure}[htp!]
	\centering
	\includegraphics[width=\textwidth]{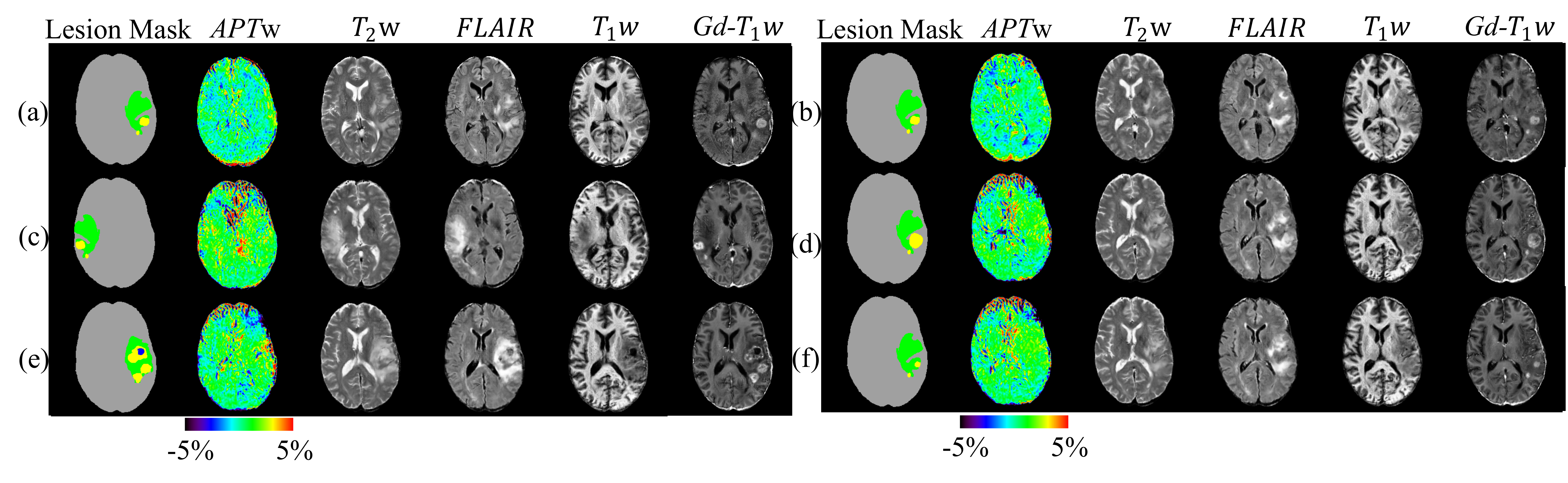}
	\vskip-15pt	\caption{Examples of lesion mask manipulations. (a) Real images. (b) Synthesized images from the original mask. (c) Synthesized images by mirroring lesion. (d) Synthesized images by increasing tumor size to 100\%. (e) Synthesized images by replacing lesion from another patient. (f) Synthesized images by shrinking tumor size to 50\%. In lesion masks, gray, green, yellow, and blue represent normal brain, edema, tumor, and cavity caused by surgery, respectively.} \label{fig4}
	\vskip-15pt	
\end{figure}
\noindent {\bf{MRI Synthesis Results.}} We evaluate the performance of different synthesis methods by qualitative comparison and human perceptual study. We compare the performance of our method with the following recent state-of-the-art synthesis methods: pix2pix \cite{au8}, pix2pixHD \cite{au9}, and Shin et al. \cite{au14}. Fig.~\ref{fig3} presents the qualitative comparison of the synthesized multi-model MRI sequences from four different methods. It can be observed that pix2pix \cite{au8} and pix2pixHD \cite{au9} fail to synthesize realistic looking human brain MR images. There is either an unreasonable brain ventricle (see the last row of Fig.~\ref{fig3}) or wrong radiographic features in the lesion region (see the fourth row of Fig.~\ref{fig3}). With the help of probability maps of white matter, gray matter and CSF, Shin et al. \cite{au14} can produce realistic brain anatomic structures for anatomic MRI sequences. However, there is an obvious disparity between the synthesized and real $APT$w sequence in both normal brain and lesion region. The boundary of the synthezied lesion is also blurred and uncertain (see red boxes in the third row of Fig.~\ref{fig3}). The proposed method produces more accurate radiographic features of lesions and more diverse anatomic structure based on the human anatomy prior provided by atlas.\\ 
\noindent {\bf{Human Perceptual Study.}} To verify the pathological information of the synthesized images, we conduct the human perceptual study by an expert neuroradiologist and the corresponding preference rates are reported in Table~\ref{tab3}. It is clear that the images generated by our method are more preferred by an expert neuroradiologist than others showing the practicality of our synthesis method.\\ 
\begin{table}[htp!]
	\centering
%	\scriptsize
	\vskip-25pt
	\setlength{\tabcolsep}{4pt}
	\caption{Preference rates corresponding to the human perceptual study.} \label{tab3}
	\vskip-5pt
	\begin{tabular}{lccccc}
		\hline
		& Real  & Our    & Shin et al. \cite{au14} & pix2pixHD \cite{au9} & pix2pix \cite{au8} \\ \hline
		& 100\% & \textbf{72.1}\% & 65.6\%      & 39.3\%    & 16.4\%  \\ \hline
	\end{tabular}
\vskip-25pt	
\end{table} 

\noindent {\bf{Data Augmentation Results.}} To further evaluate the quality of the synthezied images, we perform data augmentation by using the synthesized images in training and then perform lesion segmentation. Evaluation metrics in BraTS \cite{au3} challenge (i.e. Dice score, Hausdorff distance (95\%), Sensitivity, and Specificity) are used to measure the performance of different methods. The data augmentation by synthesis is evaluated by the improvement for lesion segmentation models. We arbitrarily control lesion information to synthesize different number of data for augmentation. The detail of mechanism for manipulating lesion mask can be found in supplementary material. To simulate the piratical usage of data augmentation, we conduct experiments in the manner of utilizing all real data. In each experiment, we vary the percentage of synthezied data to observe the contribution for data augmentation. Table~\ref{tab1} shows the calculated segmentation performance. Comparing with the baseline experiment that only uses real data, the synthesized data from pix2pix \cite{au8} and pix2pixHD  \cite{au9} degrade the segmentation performance. The performance of Shin et al. \cite{au14} improves when synthesized data is used for segmentation, but the proposed method outperforms other methods by a large margin. Fig.\ref{fig4} demonstrates the robustness of the proposed model under different lesion mask manipulations (e.g. changing the size of tumor, moving lesion location, and even reassembling lesion information between patients).  As can be seen from this figure, our method is robust to various lesion mask manipulations.\\ 
\begin{table}[ht]
	\setlength{\tabcolsep}{0.2pt}
	\centering
	\vskip-5pt
	\caption{Quantitative results corresponding to image segmentation when  the synthesized data is used for data augmentation. For each experiments, the first row reports the percentage of synthesized/real data for training and the number of instances of synthesized/real data in parentheses. Exp.4 reports the results of ablation study.}\label{tab1}
	\scriptsize
	\begin{tabular}{ccccccccccccccccc}
		\hline
		\multicolumn{17}{c}{Exp.1:  50\% Synthesized+ 50\% Real (1080 + 1080)}                                                                                                                                                                                                                                                                                                                       \\ \hline
		\multicolumn{1}{l}{} & \multicolumn{1}{l}{}  & \multicolumn{3}{c}{Dice Score}                                        & \multicolumn{1}{l}{} & \multicolumn{3}{c}{Hausdorff95 Distance}                              & \multicolumn{1}{l}{} & \multicolumn{3}{c}{Sensitivity}                                       & \multicolumn{1}{l}{} & \multicolumn{3}{c}{Specificity}                  \\ \hline
		& \multicolumn{1}{c|}{} & Edema          & Cavity         & \multicolumn{1}{c|}{Tumor}          &                      & Edema          & Cavity         & \multicolumn{1}{c|}{Tumor}          &                      & Edema          & Cavity         & \multicolumn{1}{c|}{Tumor}          &                      & Edema          & Cavity         & Tumor          \\
		pix2pix \cite{au8}              & \multicolumn{1}{c|}{} & 0.589          & 0.459          & \multicolumn{1}{c|}{0.562}          &                      & 13.180         & 21.003         & \multicolumn{1}{c|}{10.139}         &                      & 0.626          & 0.419          & \multicolumn{1}{c|}{0.567}          &                      & 0.995          & 0.998          & 0.999          \\
		pix2pix HD \cite{au9}          & \multicolumn{1}{c|}{} & 0.599          & 0.527          & \multicolumn{1}{c|}{0.571}          &                      & 17.406         & 8.606          & \multicolumn{1}{c|}{10.369}         &                      & 0.630          & 0.494          & \multicolumn{1}{c|}{0.570}          &                      & 0.996          & 0.998          & 0.999          \\
		Shin et al. \cite{au14}          & \multicolumn{1}{c|}{} & 0.731          & 0.688          & \multicolumn{1}{c|}{0.772}          &                      & 7.306          & 6.290          & \multicolumn{1}{c|}{6.294}          &                      & 0.701          & 0.662          & \multicolumn{1}{c|}{0.785}          &                      & 0.997          & 0.999          & 0.999          \\
		our                  & \multicolumn{1}{c|}{} & \textbf{0.794} & \textbf{0.813} & \multicolumn{1}{c|}{\textbf{0.821}} & \textbf{}            & \textbf{6.049} & \textbf{1.568} & \multicolumn{1}{c|}{\textbf{2.293}} & \textbf{}            & \textbf{0.789} & \textbf{0.807} & \multicolumn{1}{c|}{\textbf{0.841}} & \textbf{}            & \textbf{0.997} & \textbf{0.999} & \textbf{0.999} \\ \hline
		\multicolumn{17}{c}{Exp.2:  25\% Synthesized+ 75\% Real (540 + 1080)}                                                                                                                                                                                                                                                                                                                        \\ \hline
		pix2pix \cite{au8}             & \multicolumn{1}{c|}{} & 0.602          & 0.502          & \multicolumn{1}{c|}{0.569}          & \multirow{4}{*}{}    & 10.706         & 9.431          & \multicolumn{1}{c|}{10.147}         & \multirow{4}{*}{}    & 0.640          & 0.463          & \multicolumn{1}{c|}{0.579}          & \multirow{4}{*}{}    & 0.995          & 0.999          & 0.998          \\
		pix2pix HD \cite{au9}          & \multicolumn{1}{c|}{} & 0.634          & 0.514          & \multicolumn{1}{c|}{0.663}          &                      & 17.754         & 9.512          & \multicolumn{1}{c|}{9.061}          &                      & 0.670          & 0.472          & \multicolumn{1}{c|}{0.671}          &                      & 0.996          & 0.999          & 0.999          \\
		Shin et al. \cite{au14}          & \multicolumn{1}{c|}{} & 0.673          & 0.643          & \multicolumn{1}{c|}{0.708}          &                      & 14.835         & 7.798          & \multicolumn{1}{c|}{6.688}          &                      & 0.664          & 0.602          & \multicolumn{1}{c|}{0.733}          &                      & 0.997          & 0.999          & 0.998          \\
		our                  & \multicolumn{1}{c|}{} & \textbf{0.745} & \textbf{0.780} & \multicolumn{1}{c|}{\textbf{0.772}} &                      & \textbf{8.779} & \textbf{6.757} & \multicolumn{1}{c|}{\textbf{4.735}} &                      & \textbf{0.760} & \textbf{0.788} & \multicolumn{1}{c|}{\textbf{0.805}} &                      & \textbf{0.997} & \textbf{0.999} & \textbf{0.999} \\ \hline
		\multicolumn{17}{c}{Exp.3:  0\% Synthesized + 100\% Real (0 + 1080)}                                                                                                                                                                                                                                                                                                                         \\ \hline
		Baseline             & \multicolumn{1}{c|}{} & \textbf{0.646} & \textbf{0.613} & \multicolumn{1}{c|}{\textbf{0.673}} & \textbf{}            & \textbf{8.816} & \textbf{7.856} & \multicolumn{1}{c|}{\textbf{7.078}} & \textbf{}            & \textbf{0.661} & \textbf{0.576} & \multicolumn{1}{c|}{\textbf{0.687}} & \textbf{}            & \textbf{0.996} & \textbf{0.999} & \textbf{0.998} \\ \hline
		\multicolumn{17}{c}{Exp.4: \textbf{Ablation Study}}  \\ \hline
		w/o      Stretch-out                             & \multicolumn{1}{c|}{} & 0.684                     & 0.713                     & \multicolumn{1}{c|}{0.705}          &                      & 6.592                     & 5.059                     & \multicolumn{1}{c|}{4.002}          &                      & 0.708                     & 0.699                     & \multicolumn{1}{c|}{0.719}          &                      & 0.997                     & 0.999                     & 0.999                     \\
		w/o  Label-wise D                         & \multicolumn{1}{c|}{} & 0.753                     & 0.797                     & \multicolumn{1}{c|}{0.785}          &                      & 7.844                     & 2.570                     & \multicolumn{1}{c|}{2.719}          &                      & 0.735                     & 0.780                     & \multicolumn{1}{c|}{0.783}          &                      & 0.998                     & 0.999                     & 0.999                     \\
		w/o     Atlas                      & \multicolumn{1}{c|}{} & 0.677                     & 0.697                     & \multicolumn{1}{c|}{0.679}          &                      & 13.909                    & 11.481                    & \multicolumn{1}{c|}{7.123}          &                      & 0.691                     & 0.689                     & \multicolumn{1}{c|}{0.723}          &                      & 0.997                     & 0.999                     & 0.998                     \\
		w/o  $\mathcal{L}_{C}$ & \multicolumn{1}{l|}{} & 0.728 & 0.795 & \multicolumn{1}{c|}{0771}         & \multicolumn{1}{l}{} & 8.604 & 3.024 & \multicolumn{1}{c|}{3.233}          & \multicolumn{1}{l}{} &0.738& 0.777& \multicolumn{1}{c|}{0.777}          & \multicolumn{1}{l}{} & 0.997 & 0.999 & 0.999 \\ 
		Our       & \multicolumn{1}{c|}{} & \textbf{0.794}            & \textbf{0.813}            & \multicolumn{1}{c|}{\textbf{0.821}} & \textbf{}            & \textbf{6.049}            & \textbf{1.568}            & \multicolumn{1}{c|}{\textbf{2.293}} & \textbf{}            & \textbf{0.789}            & \textbf{0.807}            & \multicolumn{1}{c|}{\textbf{0.841}} & \textbf{}            & \textbf{0.998}            & \textbf{0.999}            & \textbf{0.999} \\    \hline
	\end{tabular}
	\vskip-15pt	
\end{table}

\noindent {\bf{Ablation Study. }}  We conduct extensive ablation study to separately evaluate the effectiveness of using stretch-out up-sampling module, label-wise discriminators, atlas, and lesion shape consistency loss $\mathcal{L}_{C}$ in our method using the same experimental setting as exp.1 in Table~\ref{tab1}. The contribution of modules for data augmentation by synthesis is reported in Table~\ref{tab1} exp.4. We find that when atlas is not used in our method, it significantly affects the synthesis quality due to the lack of human brain anatomy prior. Losing the customized reconstruction for each sequence (stretch-out up-sampling module) can also degrade the synthesis quality. Moreover, dropping either  $\mathcal{L}_{C}$ or label-wise discriminators in the training reduces the performance, since the shape consistency loss and the specific supervision on ROIs are not used to optimize the generator to produce more realistic images. 
\section{Conclusion}
We proposed an effective generation model for multi-model anatomic and molecular $APT$w  MRI sequences. It was shown that the proposed multi-task optimization under adversarial training further improves the synthesis quality in each ROI. The synthesized data can be used for data augmentation, particularly for those images with pathological information of malignant gliomas, to improve the performance of  segmentation. Moreover, the proposed approach is an automatic, low-cost solution to produce high quality data with diverse content that can be used for training of data-driven methods. \\
\indent In our future work, we will generalize data augmentation by synthesis to other tasks, such as classification.  Furthermore, the proposed method will be extended to 3D synthesis once better quality molecular MRI data is available for training the models. \\\\  
\noindent {\bf{Acknowledgment. }} Vishal M. Patel was supported by the National Science Foundation grant 1910141. This work was supported in part by grants from the National Institutes of Health (R01CA228188).

%
% ---- Bibliography ----
%
% BibTeX users should specify bibliography style 'splncs04'.
% References will then be sorted and formatted in the correct style.
%
\bibliographystyle{splncs04}
%\bibliography{references}

\end{document}